\documentclass[conference]{IEEEtran}
\IEEEoverridecommandlockouts
\usepackage{cite}
\usepackage{amsmath,amssymb,amsfonts}
\usepackage{algorithmic}
\usepackage{graphicx}
\usepackage{textcomp}
\usepackage{xcolor}

\usepackage{multirow}
\usepackage[table,xcdraw]{xcolor}
\usepackage{colortbl}
\usepackage{amsmath}
\usepackage{subcaption}

\usepackage[dvipsnames]{xcolor}
\usepackage[most]{tcolorbox}
\definecolor{prompt}{RGB}{0,90,180}
\definecolor{speech}{RGB}{180,40,40}
\definecolor{transcript}{RGB}{20,120,20}
\newtcolorbox{promptbox}{
    colback=gray!5,
    colframe=gray!60,
    boxrule=0.4pt,
    arc=2pt,
    left=4pt,
    right=4pt,
    top=4pt,
    bottom=4pt
}

\def\BibTeX{{\rm B\kern-.05em{\sc i\kern-.025em b}\kern-.08em
    T\kern-.1667em\lower.7ex\hbox{E}\kern-.125emX}}
\begin{document}

\title{Generative vs. Encoder Large Language Models for ASR Evaluation: A Comparative Study\\
}

\author{\IEEEauthorblockN{
Thibault Bañeras-Roux\textsuperscript{1},
Shashi Kumar\textsuperscript{1},
Driss Khalil\textsuperscript{1},
Sergio Burdisso\textsuperscript{1},
Petr Motlicek\textsuperscript{1}, \\
Shiran Liu\textsuperscript{1},
Mickael Rouvier\textsuperscript{2},
Jane Wottawa\textsuperscript{3},
Richard Dufour\textsuperscript{4}}
\IEEEauthorblockA{
\textsuperscript{1}Idiap Research Institute, Martigny, Switzerland
\quad
\textsuperscript{2}LIA, Avignon Université, Avignon, France
\\
\textsuperscript{3}LIUM, Le Mans Université, Le Mans, France
\quad
\textsuperscript{4}LS2N, Nantes Université, Nantes, France
}
}


\maketitle

\begin{abstract}
Automatic Speech Recognition (ASR) is typically evaluated using Word Error Rate (WER), which poorly reflects semantic similarity. While embedding-based metrics correlate better with human judgments, the respective roles of encoder and decoder-based Large Language Models (LLMs) remain underexplored. This paper presents a comparative study of both families for ASR evaluation. We analyze BERTScore and SemDist across different LLMs, layers, and pooling strategies, showing that both metrics can achieve strong correlation with human judgments when properly configured. For decoder models, we investigate generative LLMs in two settings: pairwise hypothesis selection via prompting and direct qualitative error classification. Our results show that encoder-based metrics remain highly competitive, while generative LLMs perform strongly in hypothesis comparison and improve the interpretability of ASR evaluation.
\end{abstract}

\begin{IEEEkeywords}
Automatic Speech Recognition, Large Language Models, Semantic Evaluation, BERTScore, SemDist, Human Perception
\end{IEEEkeywords}

\section{Introduction}

Automatic Speech Recognition (ASR) plays a central role in applications such as voice interfaces, transcription services, and accessibility tools. Because of its broad impact, robust evaluation methodologies are crucial for tracking progress and guiding system development. Although Word Error Rate (WER) has long been the standard metric, it remains limited by its strict matching rules and sensitivity to surface-form variations such as casing or minor lexical differences, which can obscure meaningful improvements in perceived transcription quality.

To address these shortcomings, a variety of evaluation approaches have been introduced. In particular, semantic metrics~\cite{kim2021semantic, zhangbertscore} leveraging contextualized word representations have demonstrated stronger alignment with human judgments. Methods based on encoder architectures such as BERT~\cite{devlin2019bert} and related models~\cite{roux2022qualitative, kim2021semantic, le2016better, gordeeva2021meaning} have been especially influential in capturing contextual meaning beyond surface-level text overlap. However, comparatively little attention has been given to decoder-only large language models, including GPT~\cite{radford2018improving}, Llama~\cite{touvron2023llama}, and Gemma~\cite{team2024gemma}, despite their strong general-purpose reasoning and generation capabilities~\cite{ahuja2023mega, labrak2024biomistral, naveed2025comprehensive}. This gap raises the question of how encoder and decoder-based representations compare in structured ASR evaluation settings.

In this work, we systematically investigate the role of both encoder and decoder representations for ASR evaluation along complementary dimensions. First, we compare semantic representations derived from both encoder models and decoder LLMs, studying how different pooling strategies affect their ability to encode meaningful sentence-level similarity. Second, we examine the "LLM-as-judge" paradigm, assessing whether decoder-based models can directly discriminate between competing transcription hypotheses by leveraging their contextual understanding.

Our evaluation is conducted on the HATS dataset~\cite{baneras2023hats}, which provides detailed human judgments of ASR outputs and has shown that semantic metrics correlate more strongly with perception than WER. We analyze (1) semantic similarity metrics derived from encoder versus decoder embeddings under different pooling strategies, (2) pairwise hypothesis selection, and (3) the ability of LLMs to produce qualitative error assessments. Our findings indicate that decoder-based models can rival or surpass strong encoder-based metrics, while both families yields improved robustness and interpretability in comparison with standard ASR evaluation.

\section{Related Work}
\label{sec:sota}

\subsection{From Lexical to Semantic ASR Evaluation}

Automatic Speech Recognition systems are commonly evaluated using Word Error Rate, which measures the minimum number of insertions, deletions, and substitutions required to transform a hypothesis into its reference. Owing to its simplicity and reproducibility, WER has become the de facto standard for ASR evaluation. However, numerous studies have highlighted its limitations~\cite{wang2003word,morris2004and,he2011word}. Since every token error contributes equally to the final score, WER ignores the semantic importance of individual mistakes, penalizing harmless wording changes as severely as meaning-altering errors.

These shortcomings become particularly apparent when considering downstream applications. Improvements in WER do not necessarily translate into better performance in tasks that consume ASR transcripts~\cite{favre2013automatic}, nor do they consistently reflect users' perception of transcription quality~\cite{kim2022evaluating,thennal2025advocating}. As a consequence, different evaluation metrics may rank ASR systems differently~\cite{baneras2024comprehensive}, suggesting that the choice of metric should depend on the intended application, especially when transcripts are designed for humans~\cite{kafle2017evaluating,nam2019simulation}.

\subsection{Embedding-based Semantic Metrics}

To better account for meaning preservation, several semantic evaluation metrics have been proposed. Early approaches relied on static word embeddings~\cite{le2016better}, while more recent methods exploit contextual representations produced by Transformer encoders such as BERT~\cite{devlin2019bert}. These contextual embeddings capture sentence meaning beyond lexical overlap, making them particularly suitable for evaluating ASR hypotheses.

Among the most widely used semantic metrics are BERTScore~\cite{zhangbertscore}, which compares contextual token embeddings, and SemDist~\cite{kim2021semantic}, which measures the cosine distance between sentence embeddings. Both have been shown to correlate substantially better with human judgments than WER~\cite{kim2021semantic}. Their performance, however, depends on the underlying language model and on the representation extracted from it, motivating a deeper analysis of embedding choices.

Despite these improvements, embedding-based metrics remain difficult to interpret. Unlike WER or CER, which are expressed as intuitive error rates, semantic metrics generally produce cosine similarities without an obvious semantic interpretation. Several works have therefore proposed incorporating semantic severity into lexical metrics~\cite{gordeeva2021meaning,roy2021semantic} or analysing ASR errors according to their semantic impact~\cite{baneras2024paradigm}, providing more informative evaluations than a single similarity score.

\subsection{Human-centered ASR Evaluation}

A complementary research direction aims to design evaluation metrics that better reflect human perception~\cite{itoh2015metric,sasindran2024semascore}. This has motivated the collection of datasets containing explicit human judgments of transcription quality, enabling direct comparison between automatic metrics and perceived quality.

The HATS dataset~\cite{baneras2023hats} is one such benchmark, providing pairwise human preferences over French ASR hypotheses. It demonstrated that semantic metrics achieve substantially higher agreement with human annotators than WER, confirming observations previously reported on proprietary datasets~\cite{kim2022evaluating}. More recently,~\cite{thennal2025advocating} showed that Character Error Rate (CER) can itself provide a stronger approximation of human judgments than WER in multilingual settings.

While previous work has primarily focused on encoder-based semantic representations, the emergence of decoder-based Large Language Models raises the question of whether generative models can also serve as effective ASR evaluation tools. This paper investigates both encoder and decoder LLMs, comparing their representations for semantic metrics and evaluating the ability of generative models to directly assess transcription quality.


\section{Embedding-based ASR Evaluation}

As said previously, WER poorly reflects human perception. While embedding-based metrics correlate better with human judgments, their behavior depends strongly on the underlying representations and extraction strategy. In this work, we study encoder and decoder-based LLM embeddings for semantics metrics according to human expectations.

\subsection{Evaluation Setup}

We rely on the HATS dataset as a reference for comparing automatic metrics against human judgments. Following prior work, we consider three subsets based on inter-annotator agreement: (i) 100\% agreement, (ii) $\geq$70\% agreement, and (iii) the full dataset. This allows us to analyze metric robustness under varying levels of annotation ambiguity. Agreement is computed as:
\[
\text{Agreement} = \frac{\max(A, B)}{A + B}.
\]

As a baseline, WER and CER achieve agreements of 63\% and 77\% respectively on the 100\% agreement subset. Since embedding-based metrics produce continuous similarity scores, ties do not occur in our experiments, unlike discrete lexical metrics such as WER or CER.

\subsection{BERTScore with Encoder and Decoder LLMs}

\begin{figure}[htbp]
\centering
\includegraphics[width=\linewidth]{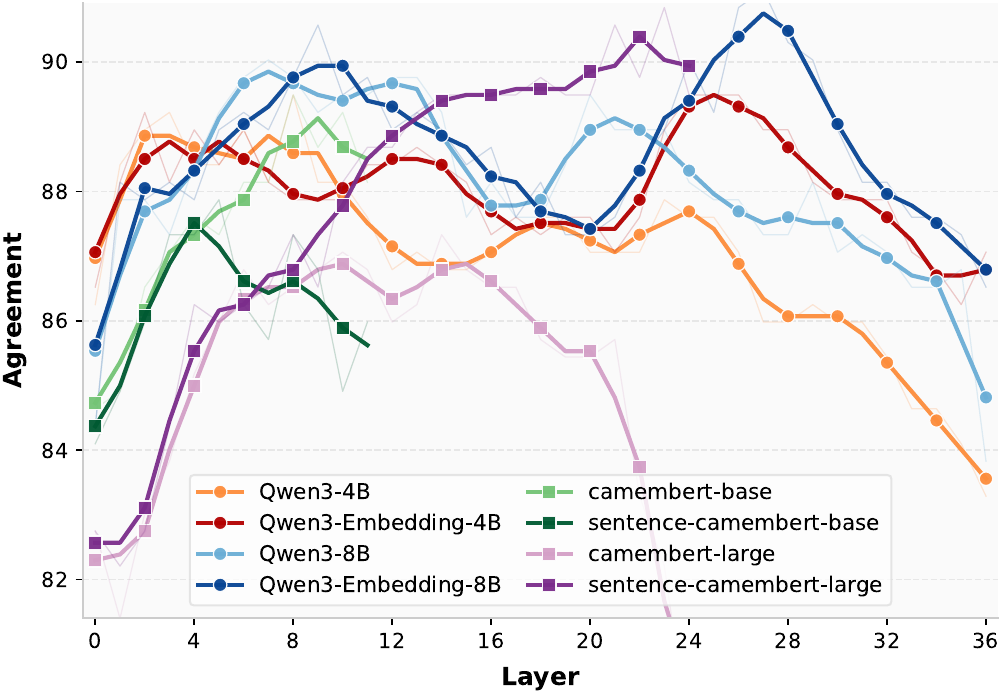}
\caption{Agreement with human judgments of BERTScore metric across models and representation layers. Human annotations are obtained from the HATS dataset subset with annotator consensus.}
\label{fig:bertscore_layer}
\end{figure}

To investigate the influence of contextual representations on semantic evaluation, we compute BERTScore using the embeddings extracted from every transformer layer of each evaluated encoder and decoder LLM. The correlation between the resulting BERTScore values and human judgments on the HATS dataset is then measured, allowing us to identify which models and representation layers best capture the semantic similarity perceived by human annotators.


Figure~\ref{fig:bertscore_layer} reports the agreement obtained for each model and layer. A first observation is that the optimal representation layer strongly depends on the model. Although intermediate layers often provide the highest agreement, the final layer remains optimal for several models, making it difficult to predict \emph{a priori} which layer should be used. Consequently, layer selection should be regarded as an important hyperparameter when applying BERTScore for ASR evaluation.

Comparing the encoder models reveals that sentence-level fine-tuning primarily affects the last layers of the network, while earlier layers remain remarkably similar to those of the corresponding CamemBERT models. This specialization can lead to substantial improvements, as illustrated by Sentence-CamemBERT-large, but it is not systematic: Sentence-CamemBERT-base has decreasing performances in comparison with CamemBERT-base, suggesting that sentence-level fine-tuning is beneficial only when the underlying representation is sufficiently expressive.

A similar behavior is observed for decoder models. The embedding-specialized Qwen3 models generally improve over the original model, with the largest differences appearing in the deepest layers. However, the last layers do not seem to be the most suitable representations for BERTScore.

Overall, both encoder and decoder-based LLMs achieve strong performance when an appropriate representation layer is selected. The best results are obtained with Sentence-CamemBERT-large and Qwen3-Embedding-8B, demonstrating that dedicated sentence or embedding-oriented fine-tuning can be benefic to both architectures. Interestingly, encoder models achieve performance comparable to the best decoder models despite having substantially fewer parameters, highlighting their efficiency for embedding-based ASR evaluation.

\begin{table}[hbt]
\centering
\small
\begin{tabular}{l@{\hspace{6pt}}r@{\hspace{5pt}}r@{\hspace{3pt}}r@{\hspace{7pt}}r@{\hspace{4pt}}r}
\hline
\textbf{Model} & \textbf{Last} & \textbf{Mean} & \textbf{Mean*} & \textbf{Wgt.} & \textbf{Wgt.*} \\
\hline
Gemma-2-2b & 73 & 79 & 79 & 79 & 78 \\
Gemma-2b & 76 & 80 & 80 & 79 & 80 \\
Gemma-3-1b-pt & 75 & 76 & 74 & 77 & 76 \\
Gemma-3-27b-it & 77 & 83 & 83 & 81 & 82 \\
Gemma-4-31B & 72 & 75 & 73 & 77 & 76 \\
Gemma-7b & 65 & 69 & 71 & 71 & 72 \\
Qwen3-0.6B & 70 & 80 & 80 & 78 & 78 \\
Qwen3-0.6B-Base & 74 & 79 & 79 & 81 & 80 \\
Qwen3-1.7B & 72 & 83 & 83 & 80 & 81 \\
Qwen3-1.7B-Base & 74 & 82 & 81 & 81 & 82 \\
Qwen3-30B & 72 & 84 & 84 & 83 & 82 \\
Qwen3-4B & 77 & 85 & 83 & 82 & 81 \\
Qwen3-4B-Base & 74 & 82 & 83 & 82 & 82 \\
Qwen3-8B & 75 & 85 & 83 & 83 & 83 \\
Qwen3-8B-Base & 71 & 78 & 78 & 80 & 77 \\
Qwen3-Embedding-0.6B & \textbf{88} & 84 & 85 & 85 & 84 \\
Qwen3-Embedding-4B & \textbf{88} & 86 & 85 & 86 & 84 \\
Qwen3-Embedding-8B & \textbf{88} & \textbf{89} & \textbf{87} & \textbf{89} & \textbf{87} \\
Qwen3.5-27B & 59 & 77 & 76 & 74 & 76 \\
Qwen3.5-35B & 68 & 68 & 68 & 67 & 68 \\
CamemBERT-Base & 83 & 84 & 84 & 83 & 83 \\
CamemBERT-Large & 74 & 82 & 81 & 80 & 80 \\
Sentence-CamemBERT-Base & 84 & 86 & 86 & 85 & 85 \\
Sentence-CamemBERT-Large & \textbf{89} & \textbf{90} & \textbf{89} & \textbf{88} & \textbf{88} \\
\hline
Mean & \textit{74} & \textit{80} & \textit{80} & \textit{80} & \textit{80} \\
\hline
\end{tabular}
\caption{Agreement with human judgments of SemDist with different models and pooling strategies, using last-layer representations. Human annotation are obtained from the HATS dataset subset with annotator consensus.}
\label{tab:semdist_comparison}
\end{table}

\subsection{SemDist with Encoder and Decoder LLMs}

We evaluate semantic similarity using the SemDist metric, defined as a distance derived from cosine similarity between sentence-level embeddings obtained by aggregating token representations from LLMs.


Given a sequence of token embeddings produced by an LLM, a fixed-size sentence representation is obtained via pooling. We evaluate multiple strategies, where $t_i$ denotes the embedding of token $i$ in a sequence of length $n$:

\begin{itemize}
    \item \textbf{Last token}: $t_n$
    \item \textbf{Mean}: $\frac{1}{n}\sum_{i=1}^{n} t_i$
    \item \textbf{Mean without last}: $\frac{1}{n-1}\sum_{i=1}^{n-1} t_i$
    \item \textbf{Weighted mean}: $\frac{\sum_{i=1}^{n} i\, t_i}{\sum_{i=1}^{n} i}$
    \item \textbf{Weighted mean without last}: $\frac{\sum_{i=1}^{n-1} i\, t_i}{\sum_{i=1}^{n-1} i}$
\end{itemize}


We evaluate all combinations of models, layers, and pooling strategies in terms of correlation with human annotations on the HATS dataset. Table~\ref{tab:semdist_comparison} summarizes performance across models and pooling methods.

Overall, the results indicate that model size alone is not a reliable predictor of performance. While larger models can be beneficial in some cases, smaller or mid-sized models can achieve comparable or even better correlations depending on the layer and pooling strategy. This suggests that the quality of semantic representations is driven more by architectural and training choices than by scale alone.

A second observation concerns pooling strategies. Excluding embedding-specialized models, last-token representations are generally suboptimal. This is consistent with the fact that autoregressive models are trained to predict the next token rather than to encode global sentence meaning in the final hidden state. In contrast, mean and weighted pooling are consistently strong, suggesting that aggregating token-level information provides a more stable estimate of sentence-level semantics, especially when hypothesis and reference lengths are comparable.

As expected, embedding-specialized models behave differently: particularly for Qwen models, the last-token representation becomes significantly more competitive, often matching or surpassing alternative pooling strategies. This confirms that embedding-oriented fine-tuning can reshape the role of the final hidden state, making it more suitable as a global semantic representation.

\begin{figure}[htbp]
\centering
\includegraphics[width=\linewidth]{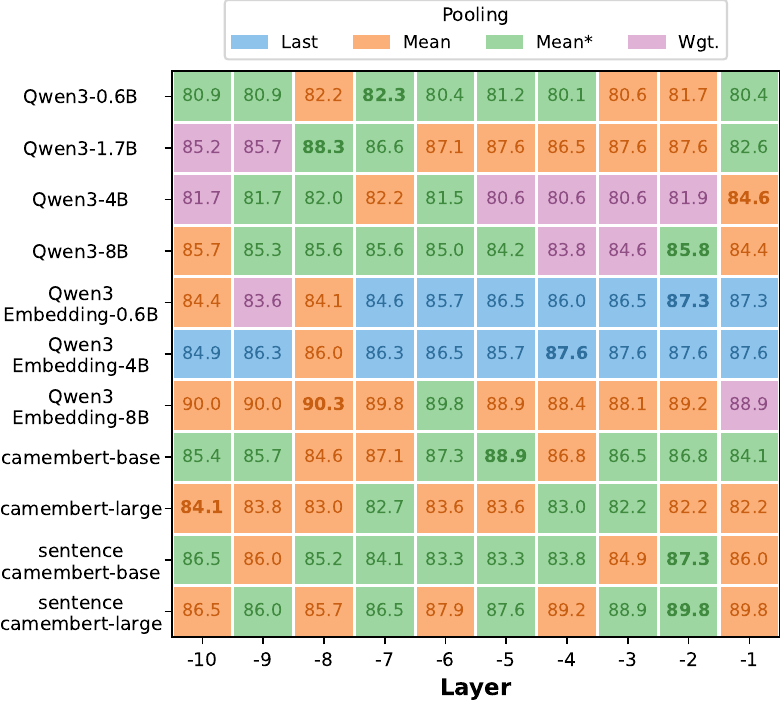}
\caption{Agreement of human judgments of SemDist across encoder and decoder LLMs as a function of layer depth (from the 10th-to-last layer to the last layer). Each cell reports the best pooling strategy. Human annotation are obtained from the HATS dataset subset with annotator consensus.}
\label{fig:semdist_heatmap}
\end{figure}

The results in Table~\ref{tab:semdist_comparison} focus on the impact of model choice and pooling strategies when using representations from the last transformer layer. This allows us to isolate how different aggregation methods influence SemDist performance under a fixed representation depth. In contrast, Figure~\ref{fig:semdist_heatmap} extends this analysis by varying the representation layer, enabling a more comprehensive view of how both pooling strategy and layer depth jointly affect performance across encoder and decoder LLMs.

We observe that the optimal layer is highly model-dependent: neither early nor final layers are universally optimal, and intermediate layers are not systematically superior. This makes layer selection a critical design choice when using LLM embeddings for semantic evaluation.

When comparing architectures, both encoder and decoder-based models can achieve strong performance under appropriate layer and pooling configurations. Encoder models such as Sentence-CamemBERT-large achieve consistently high performance with relatively compact architectures, while decoder-based embedding models such as Qwen3-Embedding-8B reach similar or slightly higher scores. However, this comes at a significantly higher parameter cost in many cases, suggesting that encoders remain highly competitive in terms of efficiency.

The best overall results are obtained with Sentence-CamemBERT-large and Qwen3-Embedding-8B. Interestingly, while these models are consistently strong, the ranking of intermediate models is not stable across pooling strategies or layers. For instance, the relative ordering of Qwen variants can differ between SemDist and BERTScore, indicating that different metrics emphasize different aspects of the learned representations.

As suggested in Table~\ref{tab:semdist_comparison}, Sentence-CamemBERT-Large appears to achieve its best performance in the deeper layers, where its representations are more aligned with semantic similarity. In contrast, Qwen3-Embedding-8B exhibits a more stable behavior across layers, with consistently strong results that are less sensitive to the specific depth of representation. This suggests that Sentence-CamemBERT-Large benefits more from layer selection, while Qwen3-Embedding-8B provides more robust embeddings throughout the network.


\section{LLM as a Judge for ASR Evaluation}

Unlike the previous section, which relied on embeddings extracted from encoder and decoder LLMs to build semantic similarity metrics, we now investigate the native generative capabilities of decoder LLMs. Rather than comparing vector representations, the models are prompted directly to reason about transcription quality. We consider two complementary settings: selecting the best hypothesis between two candidates and assigning an interpretable quality label to a single hypothesis.

LLM inferences are performed using the SDialog toolkit~\cite{burdisso2026sdialog}.

\subsection{Pairwise Hypothesis Selection}
\label{sec:hypA-or-hypB}

The first experiment evaluates whether a generative LLM can act as a judge by selecting, given a reference transcription and two ASR hypotheses, the hypothesis that best matches human preference. Unlike embedding-based metrics, this formulation allows the model to jointly consider lexical accuracy, semantic fidelity, grammatical coherence and contextual information before making a decision.

The model is prompted in a one-shot setting consisting of a single annotated example followed by the pair to evaluate. It is asked to briefly justify its decision before ending its answer with either \texttt{A} or \texttt{B}, allowing automatic extraction of the final prediction while preserving the model's reasoning process. The complete prompt is provided in Appendix~\ref{appendix:pairwise-prompt}.

Table~\ref{tab:correlation-hypA-or-hypB} reports the agreement between LLM decisions and human preferences. Overall, performance tends to improve with newer and larger models, although model size alone does not explain the results. For example, Qwen3-8B outperforms Gemma3-27B despite having substantially fewer parameters, while Qwen3.5-27B surpasses Qwen3-30B. The best results are obtained by GPT-4.1 (94\%) and the open-weight Qwen3.5-35B (92\%) on the subset with total consensus, showing that open models can approach proprietary systems on this task.

Previously, as shown in Table~\ref{tab:semdist_comparison}, we observed that Qwen3-1.7B produces good representations for a semantic metric such as SemDist, while Qwen3.5-35B does not appear to achieve comparable performance in this setting. Interestingly, we observe the opposite trend in the pairwise hypothesis selection task, where Qwen3.5-35B significantly outperforms Qwen3-1.7B. This contrast suggests that the ability to produce useful continuous embeddings is not necessarily aligned with the ability to perform discrete comparative reasoning. One possible explanation is that intrinsic evaluation of a single hypothesis relies on different capabilities than step-by-step comparative reasoning between two hypotheses.

For this specific task, these results also suggest that generative LLM outperform traditional lexical metrics such as WER and CER, as well as the best embedding-based semantic metrics evaluated in the previous section. Unlike these metrics, which compare hypotheses through fixed similarity functions, generative LLMs directly reason over complete transcriptions before making a decision. This suggests that they capture aspects of transcription quality that are not necessarily reflected in semantic representation, such as grammatical coherence, semantic consistency or tolerance to minor disfluencies.

Furthermore, its high agreement with human judgments suggests that generative LLMs could reduce the amount of manual annotation required when constructing perceptual evaluation datasets, thereby lowering annotation costs.

\begin{table}[htb]
\centering
\begin{tabular}{l@{\hspace{25pt}}c@{\hspace{20pt}}c@{\hspace{30pt}}c}
\hline
\textbf{LLM} & \textbf{$=$ 100 \%} & \textbf{$\geq$ 70 \%} & \textbf{Full} \\ \hline
GPT-4.1 & \textbf{94} & \textbf{85} & \textbf{79} \\ 
Gemma3-27B & 72 & 63 & 61 \\ 
Gemma4-31B & 87 & 78 & 73 \\
Qwen3.5-35B & \textbf{92} & \textbf{83} & \textbf{78} \\
Qwen3.5-27B & 91 & 83 & 77 \\
Qwen3-30B & 84 & 75 & 71 \\
Qwen3-8B & 80 & 74 & 72 \\ 
Qwen3-1.7B & 59 & 58 & 56 \\
Qwen3-0.6B & 50 & 47 & 47 \\ \hline
\end{tabular}%
\caption{Agreement of human judgments (HATS) with LLM choices (selecting the best hypothesis), with respect to annotator agreement level.}
\label{tab:correlation-hypA-or-hypB}
\end{table}


\subsection{Direct Classification of Hypotheses}
\label{sec:llm-classify-category}
 
 
 
 
In this section, we evaluate the capacity of language models to classify a hypothesis given a reference. 
Unlike the previous experiment, each $(\text{reference},\text{hypothesis})$ pair is evaluated independently. The objective is to determine whether an LLM can provide an absolute and interpretable assessment of transcription quality without requiring another hypothesis for comparison.

Given a reference and a hypothesis, the model must assign one of the following labels (see prompt in Appendix~\ref{appendix:direct-classification}):
\begin{itemize}
    \item \textbf{identical}: the hypothesis is identical to the reference (or differ only in case or hyphenation);
    \item \textbf{useful}: meaning is preserved despite minor errors (normalization, punctuation, spelling, capitals, abbreviations, slight syntactic variations without loss of comprehension);
    \item \textbf{bad}: meaning is partially altered (significant errors on keywords, important substitutions or omissions, but some content remains understandable);
    \item \textbf{incomprehensible}: meaning is completely lost (the sentence cannot be understood or correctly interpreted relative to the reference).
\end{itemize}

This formulation also enables a new way of reporting ASR performance. Instead of relying solely on a scalar metric such as WER or BERTScore, an LLM can assign a qualitative label to every transcription, yielding a distribution of qualitative outputs. Such a distribution is considerably more interpretable, as it characterizes the nature and severity of transcription errors rather than reducing system performance to a single score.

Since the HATS dataset does not contain such qualitative labels, direct evaluation against human annotations is not possible. Instead, we compare the predicted categories with SemDist scores computed using one the best-performing embedding model identified previously, Sentence-CamemBERT-large. This provides an independent semantic similarity signal that is known to correlate well with human perception.

Table~\ref{tab:llm-category-correlation} reports both Spearman and Pearson correlations between the predicted categories and SemDist. Negative correlations are expected because lower SemDist values correspond to better transcriptions, whereas the predicted categories are ordered from \emph{identical} (3) to \emph{incomprehensible} (0). Larger LLMs generally exhibit stronger correlations, with GPT-4.1 achieving the best overall results.

 

\begin{table}[]
\centering
\begin{tabular}{l@{\hspace{30pt}}r@{\hspace{30pt}}r}
\hline
\textbf{LLM} & \textbf{Spearman} & \textbf{Pearson} \\ \hline
GPT-4.1 & -0.66 & -0.63 \\
Gemma4-31B & -0.63 & -0.63 \\
Qwen3-30B & -0.63 & -0.66 \\
Qwen3.5-27B & -0.60 & -0.61 \\
Qwen3.5-35B & -0.58 & -0.55 \\
Qwen3-1.7B & -0.46 & -0.50 \\
Qwen3-0.6B & -0.18 & -0.22 \\ \hline
\end{tabular}%
\caption{Correlation between generated categories and SemDist with embeddings from the last layer of Sentence-CamemBERT-Large using mean pooling.}
\label{tab:llm-category-correlation}
\end{table}

 

\begin{figure}[htbp]
\centering
\includegraphics[width=\linewidth]{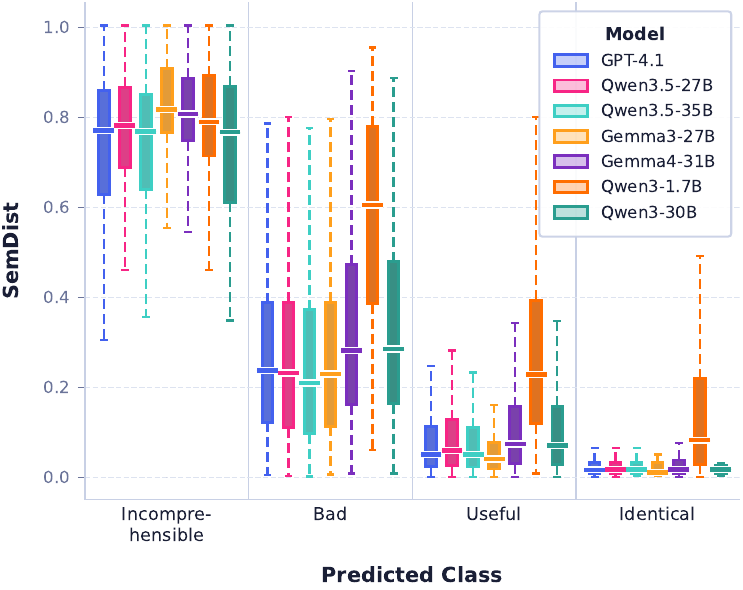}
\caption{Distribution of SemDist scores across LLM-predicted quality classes.}
\label{fig:model_pooling_performance}
\end{figure}

Figure~\ref{fig:model_pooling_performance} further illustrates the distribution of SemDist values for each predicted category. The four classes follow the expected ordering, with \emph{identical} hypotheses associated with the highest semantic similarity and \emph{incomprehensible} hypotheses with the lowest. However, substantial overlap remains between adjacent categories, particularly between \emph{useful} and \emph{bad}, explaining the moderate correlation coefficients observed in Table~\ref{tab:llm-category-correlation}.

Although these results are not yet strong enough to replace continuous semantic metrics, they demonstrate that generative LLMs can provide meaningful qualitative descriptions of ASR errors. Such descriptions offer an important interpretability advantage over scalar metrics such as WER or SemDist and constitute a promising direction for future ASR evaluation. Improving the consistency of these qualitative judgments, potentially through better prompting, dedicated fine-tuning or hybrid evaluation strategies, remains an interesting avenue for future work.


\section{Conclusion}

This paper systematically evaluates the use of large language models for ASR evaluation through three complementary perspectives.

First, we study embedding-based semantic metrics derived from both encoder and decoder LLMs. We show that BERTScore and SemDist can achieve strong agreement with human judgments when appropriate combinations of model, layer, and pooling strategy are selected. Performance does not depend solely on model scale and no single layer consistently dominates across architectures. While decoder models have slightly superior performance, encoder-based LLMs remain highly competitive, despite typically having lower computational cost.

Second, we investigate generative LLMs as direct evaluators through a pairwise hypothesis selection task. We find that state-of-the-art models achieve very high agreement with human preferences, outperforming both classical metrics such as WER and CER and embedding-based semantic metrics. These results indicate that generative models can serve as strong relative judges of ASR system outputs, capturing aspects of transcription quality that go beyond lexical similarity.

Third, we explore the use of LLMs for qualitative classification of transcription errors. While correlations with continuous semantic scores remain moderate, the results show consistent ordering of error severity and suggest that LLMs are promising solutions to provide interpretable structured feedback on ASR outputs. In particular, they enable the construction of distributions over qualitative error categories, offering a more detailed view of system behavior than single-score metrics.

Overall, our results highlight that LLMs provide a flexible framework for ASR evaluation. Encoder-based representations yield strong and efficient semantic metrics, while generative models enable both accurate comparative judgments and interpretable qualitative analysis. Taken together, these approaches open the way toward more human-aligned and interpretable evaluation methodologies for speech recognition systems.

\appendix

\section{Prompts Used for LLM-based Evaluation}

All experiments rely on fixed instruction-style prompts. This appendix reports the exact prompts used in the experiments described in Section~\ref{sec:hypA-or-hypB} and Section~\ref{sec:llm-classify-category}. The prompts are provided exactly as used during inference (translated from French for readability), without any modification or prompt engineering beyond the in-context example.

\subsection{Prompt for Pairwise Hypothesis Selection}
\label{appendix:pairwise-prompt}

The LLM input follows the template:

\begin{promptbox}
\footnotesize\fontfamily{zi4}\selectfont
\textcolor{prompt}{<s>USER: A reference is an exact transcription of audio. Two incorrect hypotheses are proposed.\\
Explain your reasoning and end your sentence by writing 'A', 'B', 'a', or 'b' if undecided.}\\[0.3em]

\textcolor{prompt}{Reference:} \textcolor{transcript}{it's to him self}\\
\textcolor{prompt}{Hypothesis A:} \textcolor{transcript}{are to him self}\\
\textcolor{prompt}{Hypothesis B:} \textcolor{transcript}{it's uh to him-self}\\[0.6em]

\textcolor{prompt}{ASSISTANT:} \textcolor{transcript}{Although hypothesis B contains a disfluency ('uh'), it corresponds much better to the reference in terms of words and meaning. The disfluency can be tolerated if it is part of the original, while hypothesis A's grammatical error is more problematic. Therefore, the most acceptable transcription is hypothesis B.}\\[0.6em]

\textcolor{prompt}{USER: Reference:} \textcolor{transcript}{\{reference\}}\\
\textcolor{prompt}{Hypothesis A:} \textcolor{transcript}{\{hypothesis A\}}\\
\textcolor{prompt}{Hypothesis B:} \textcolor{transcript}{\{hypothesis B\}}\\[0.4em]

\textcolor{prompt}{ASSISTANT:}
\end{promptbox}

\noindent
where \textcolor{transcript}{\texttt{\{reference\}}}, \textcolor{transcript}{\texttt{\{hypothesis A\}}}, and \textcolor{transcript}{\texttt{\{hypothesis B\}}} correspond to the input reference and ASR hypotheses.

\subsection{Prompt for Direct Hypothesis Classification}
\label{appendix:direct-classification}

The LLM input follows the template:

\begin{promptbox}
\footnotesize\fontfamily{zi4}\selectfont
\textcolor{prompt}{<s>USER: You are an expert evaluator of Automatic Speech Recognition (ASR). Given a reference transcription and an ASR hypothesis, classify the hypothesis into exactly one of the following categories:}\\[0.3em]

\textcolor{prompt}{IDENTICAL: the hypothesis is identical to the reference (or differs only in case or hyphenation).}\\
\textcolor{prompt}{USEFUL: the meaning is preserved despite minor errors such as normalization, punctuation, spelling, capitalization, abbreviations, or slight syntactic variations without loss of comprehension.}\\
\textcolor{prompt}{BAD: the meaning is partially altered due to significant substitutions, omissions, or errors affecting important words, but the sentence remains partially understandable.}\\
\textcolor{prompt}{INCOMPREHENSIBLE: the meaning is completely lost and the hypothesis cannot be correctly understood with respect to the reference.}\\[0.6em]

\textcolor{prompt}{Respond only with a valid JSON object following this schema:}\\
\textcolor{prompt}{\{"category": "<IDENTICAL|USEFUL|BAD|INCOMPREHENSIBLE>"\}}\\[0.8em]

\textcolor{prompt}{Reference:} \textcolor{transcript}{it's to him self}\\
\textcolor{prompt}{Hypothesis:} \textcolor{transcript}{it's uh to him-self}\\[0.8em]

\textcolor{prompt}{ASSISTANT:} \textcolor{transcript}{\{"category": "USEFUL"\}}\\[0.8em]

\textcolor{prompt}{USER: Reference:} \textcolor{transcript}{\{reference\}}\\
\textcolor{prompt}{Hypothesis:} \textcolor{transcript}{\{hypothesis\}}\\[0.5em]

\textcolor{prompt}{ASSISTANT:}
\end{promptbox}

\noindent
where \textcolor{transcript}{\texttt{\{reference\}}} and \textcolor{transcript}{\texttt{\{hypothesis\}}} denote the input transcription pair used for evaluation.

\newpage

\bibliographystyle{IEEEtran}
\bibliography{custom}


\end{document}